\documentclass[journal]{IEEEtran}
\usepackage{graphicx}
\usepackage{amsmath}
\usepackage{amssymb}
\usepackage{booktabs}
\usepackage[sort&compress]{natbib}

\usepackage{tikz}
\usepackage{amssymb}
\usepackage[utf8]{inputenc} 
\usepackage[T1]{fontenc}    
\usepackage{hyperref}       
\usepackage{url}            
\usepackage{booktabs}       
\usepackage{amsfonts}       
\usepackage{nicefrac}       
\usepackage{microtype}      
\usepackage[table]{xcolor}
\usepackage{tikz}
\usepackage{xcolor}         
\usepackage{tabularx}
\usepackage{makecell}
\usepackage{utfsym}
\usepackage{stfloats}
\usepackage{orcidlink}

\usepackage{bm}
\usepackage{url}            
\usepackage{amsfonts,amsmath}       
\usepackage{nicefrac}       
\usepackage{microtype}      
\usepackage{xcolor}
\usepackage{tabu}
\usepackage{multirow}       
\usepackage{graphicx}
\usepackage{graphics}
\usepackage{tabularx}
\usepackage{makecell}
\usepackage{caption}
\usepackage{wrapfig}
\usepackage{enumitem}
\usepackage{subcaption}
\usepackage{algorithm}
\usepackage{arydshln}
\usepackage{algpseudocode}
\usepackage{color, colortbl}

\usepackage{listings}
\definecolor{Gray}{gray}{0.95}
\definecolor{orange}{rgb}{0.9,0.5,0}

\usepackage{pifont}
\usepackage[capitalize]{cleveref}
\crefname{section}{Sec.}{Secs.}
\Crefname{section}{Section}{Sections}
\Crefname{table}{Table}{Tables}
\crefname{table}{Tab.}{Tabs.}
\begin{document}
	\title{Frequency Decoupled Framework for Screen Content Image Super-Resolution}
    \author{Xufei Wang~\orcidlink{0009-0006-8442-9526}, Qicheng Zhang~\orcidlink{0009-0004-8537-1720}, Qi Wu, Ziyang Gu and Shizhuang Weng~\orcidlink{0000-0002-7147-8496},~\IEEEmembership{Member, IEEE}

	\thanks{This work is supported in part by the National Natural Science Foundation of China under Grant 32572205 and Grant 32401708 , in part by Key Research and Development Program of Anhui Province under Grant 2023n06020017.
		
	Xufei Wang, Qicheng Zhang, Qi Wu, Ziyang Gu and Shizhuang Weng are with the School of Electronic and Information Engineering, Anhui University, Hefei 230039, China (email: wangxf024@gmail.com; setoutcheung@outlook.com; wuqi505@outlook.com; guziyang\_2025@foxmail.com; weng\_1989@126.com)
	}}
	
	\maketitle

	\begin{abstract}Methods based on implicit neural representations have demonstrated superior performance in Screen Content Image Super-Resolution (SCISR) . However, they overlooked the inherent frequency characteristics, leading to suboptimal performance. We propose a frequency decoupled framework (FDF) that rethinks SCISR from a phasor perspective by capturing structured energy in amplitude and relational continuity in phase, and jointly exploiting them with bespoke implicit representations to faithfully recover the regular textures and global configuration of Screen Content Image (SCI).
	Amplitude-Phase Factorization Network (APFN) first separates images into amplitude and phase streams, where Amplitude Clustering Module (ACM) organizes sparse yet high-energy amplitude responses into representative prototypes for periodic pattern extraction, while Phase Consistency Self-Attention (PCSA) progressively reinforces configuration through continuous consistency propagation.
    And Oscillation-Anharmonic Implicit Fitting Network (OAIF-Net) integrates periodic and coherent implicit representations for efficient exploitation of the periodic patterns and coherent context embedded in SCI. 
    Experimental results show FDF achieves state-of-the-art SCISR performance at multiple scales across four public SCI datasets. Ablation experiments further demonstrate the effectiveness of each component in extracting and exploiting periodic patterns and coherent context.

	\end{abstract}

	\section{Introduction}
\label{sec:intro}
\IEEEPARstart{W}{ith} the rapid development of multimedia applications and hardware devices, screen content images (SCI) are being used more and more widely in fields, such as remote work, cloud gaming, and online education \cite{tmm2}. There is an urgent need in the real world to enhance low-resolution (LR) image to high-resolution (HR) image. Further, Screen Content Image Super-Resolution (SCISR) has attracted great interest from researchers in recent years \cite{itsrn0}, \cite{btc}, \cite{itsrn}.
For arbitrary-scale super-resolution, LIIF \cite{liif} leverages implicit neural representation (INR), LTE \cite{lte} incorporates local texture estimation into INR, and an implicit Transformer is constructed by further introducing Transformer into INR \cite{clit}.
In SCISR, an INR-based network was proposed \cite{itsrn0}, B-spline basis functions were introduced to alleviate the Gibbs phenomenon \cite{btc}, and an implicit Transformer was introduced \cite{itsrn}.
The aforementioned SCISR method relies heavily on spatial domain and gradient-based pixel statistics to distinguish SCI from natural images (NI), while ignoring their inherent frequency characteristics, resulting in unsatisfactory performance.

\begin{figure}[t]
	\centering
	\begin{subfigure}{1\linewidth}
		\centering
		\includegraphics[width=1\linewidth]{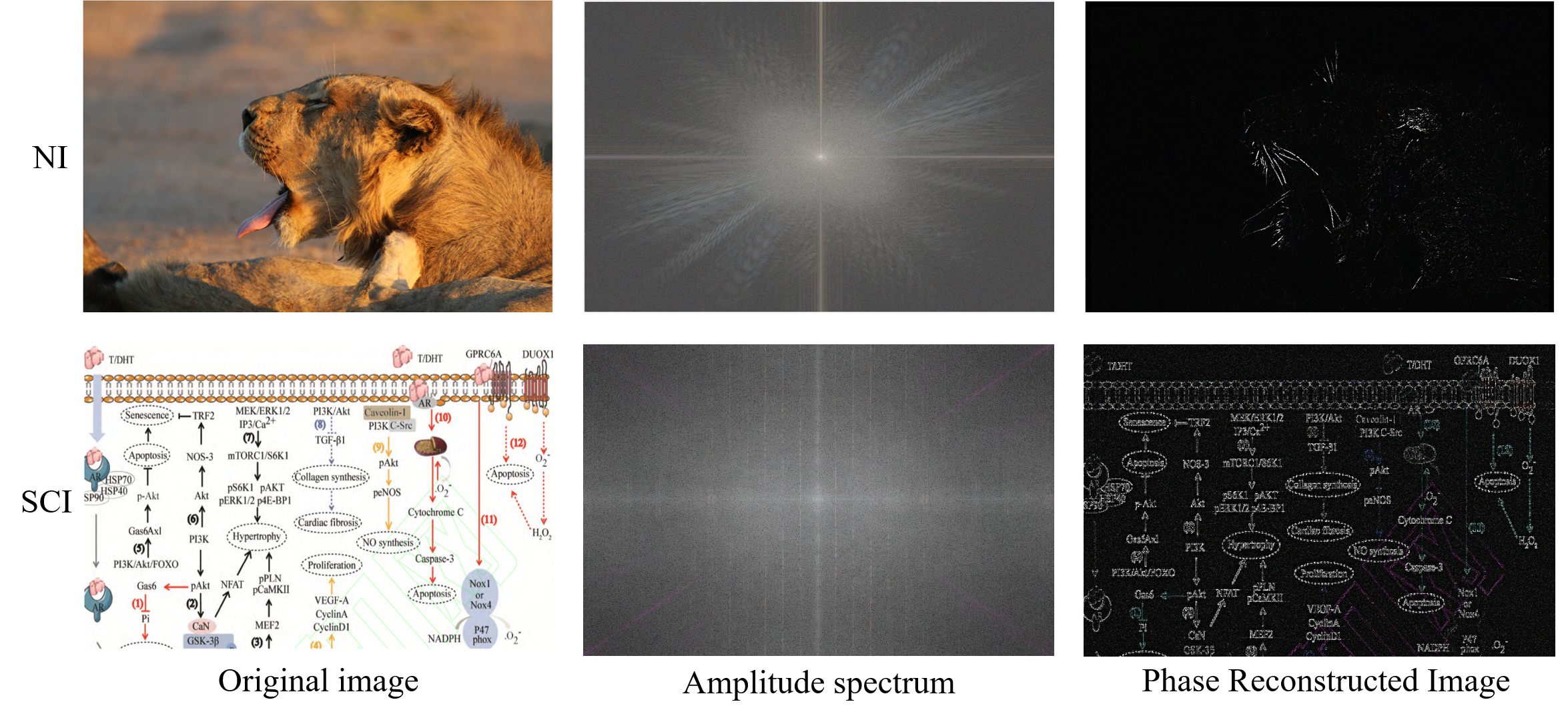}
	\end{subfigure}
	\caption{Frequency comparison between NI and SCI. From left to right are the original image, the pseudo-colored magnitude spectrum, and the phase-only reconstruction. We applied the same post-processing to the phase-reconstructed images to enhance contrast.}
	\label{SCI1}
\end{figure}
\begin{figure}[t]
	\centering
	\begin{subfigure}{1\linewidth}
		\centering
		\includegraphics[width=1\linewidth]{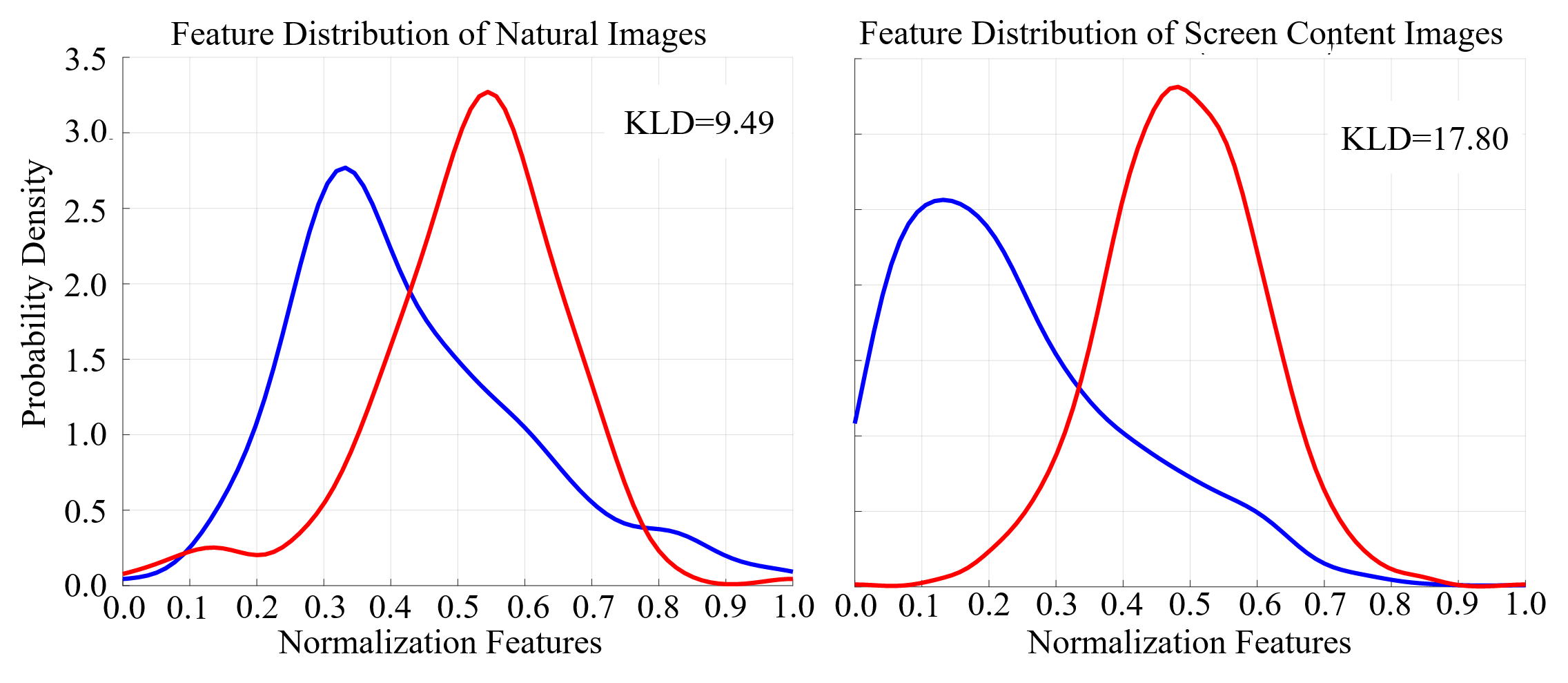}
	\end{subfigure}
	\caption{Comparison of frequency domain magnitude and phase feature distributions between NI and SCI. The left panel shows the feature distributions of natural images, while the right panel corresponds to screen images. The blue curves represent magnitude feature distributions, and the red curves represent phase feature distributions. }
	\label{SCI2}
\end{figure}

Recent advances in general vision research have highlighted the effectiveness of decoupling amplitude and phase in frequency field. Shi et al. \cite{decouple1} explored waveform-level representations of amplitude-phase separated signals to dynamically extract features from complementary spectral features; Becker et al. \cite{Thera} combined amplitude attenuation on frequency components with phase-aware interactions in activation models to capture point spread relationships, achieving accurate pixel-level correspondences. 
These studies show that modeling amplitude-phase interactions effectively utilizes frequency properties.

To investigate the intrinsic frequency characteristics of SCI, we conduct qualitative and quantitative comparisons between natural images and screen content images in the frequency domain. As illustrated in Fig.~\ref{SCI1}, the amplitude spectrum of SCI exhibits sparse yet highly structured energy distributions with numerous discrete high-energy peaks and stripe-like responses caused by repetitive layouts, sharp boundaries, and periodic textures, which is consistent with recent studies showing that embedded text, graphical elements, and repetitive structures endow SCI with frequency characteristics fundamentally different from those of NIs \cite{jiang2025fd}. Meanwhile, the phase reconstruction results preserve clearer configurational contours, stronger semantic readability, and higher global coherence. Furthermore, existing studies have characterized SCI frequency components as inherently heterogeneous \cite{omrnet}. Building upon this observation, Fig.~\ref{SCI2} presents the normalized amplitude and phase distributions modeled via kernel density estimation, where the Kullback-Leibler Divergence (KLD) between amplitude and phase in SCI is significantly larger than that in NI, indicating a stronger functional decoupling between spectral energy and structural representations. This suggests that the frequency heterogeneity of SCI manifests as a more pronounced discrepancy between amplitude-dominated energy distributions and phase-dominated structural configurations. Consequently, directly applying convolution or dense self-attention to mixed-frequency representations may introduce excessive low-energy interference. This observation is consistent with recent frequency-aware attention studies showing that uniform attention mechanisms are limited in distinguishing heterogeneous frequency components and may lead to suboptimal feature interactions \cite{dai2024freqformer}. In contrast, the phase distribution preserves rich and continuous configurations that naturally favor attention-based context modeling. These findings motivate us to explicitly decouple amplitude and phase representations and design dedicated interaction mechanisms tailored to their distinct functional characteristics. Therefore, amplitude representations are processed through structured clustering and prototype aggregation to enhance representative high-energy patterns, whereas phase representations are modeled using consistency-aware self-attention to progressively reinforce coherent contextual dependencies.

Herein, we propose a Frequency Decoupled Framework (FDF) that performs bespoke processing of the amplitude and phase information of SCI. FDF comprises an Amplitude‑Phase Factorization Network (APFN) and Oscillation-Anharmonic Implicit Fitting Network (OAIF-Net); APFN consists of an Amplitude Clustering Module (ACM) and a Phase Consistency Self‑Attention Module (PCSA).

The main contributions are summarized as:

\begin{itemize}
	\item[$\bullet$] ACM aggregates sparse yet high-energy amplitude responses into representative prototypes for enabling effective extraction of periodic pattern.
	\item[$\bullet$] PCSA progressively reinforces structurally phase dependencies through consistency-aware attention propagation to capture coherent context.
	\item[$\bullet$] OAIF-Net faithfully facilitates the exploitation of periodic pattern and coherent context during upsampling by jointly matching the corresponding implicit frequency representations.
\end{itemize}

	\section{Related Work}
\label{sec:relate}
\subsection{Single Image Super-Resolution}

Single Image Super-Resolution (SISR) is a fundamental low-level vision task that aims to reconstruct HR images from their LR counterparts \cite{tmm3}. 
SISR aims to reconstruct HR images from LR inputs and has achieved substantial progress with deep learning. SRCNN~\cite{srcnn} first introduced CNNs into SISR, followed by representative CNN-based methods such as EDSR~\cite{edsr} and RDN~\cite{rdn}, which significantly improved reconstruction performance. More recently, Transformer-based architectures including SwinIR~\cite{swinir}, SRFormer~\cite{srformer}, and HiT-SR~\cite{hit} further enhanced SR quality by modeling long-range and multi-scale dependencies through self-attention mechanisms, while LFESR~\cite{lfesr} and RIB~\cite{rib} improved fine-grained representation and
computational efficiency via local feature enhancement and efficient attention designs. Meanwhile, Arbitrary-Scale Super-Resolution (ASSR) has attracted increasing attention due to its flexibility across continuous scaling factors \cite{tmm4}. MetaSR~\cite{metasr} introduced meta-learning for arbitrary-scale reconstruction, and subsequent implicit neural representation (INR)-based methods further advanced ASSR by modeling continuous spatial mappings. LIIF~\cite{liif} employed continuous coordinate-based feature querying for RGB prediction, LTE~\cite{lte} incorporated local texture estimation to enhance detail representation, and implicit Transformer-based methods~\cite{ciaosr}, \cite{clit} further integrated contextual modeling into INR frameworks for improved arbitrary-scale reconstruction.
The confinement of these methods to natural imagery acquired by cameras in physical environments substantially limits their applicability to SCI.

\subsection{Screen Content Image Super-Resolution}

SCI consists of computer‑rendered content and typically contains periodic textures uncommon in natural images, such as sharp edges and lines \cite{tmm1}. Liu et al. \cite{hevc} first proposed an efficient SCI coding scheme by adapting processing methods originally designed for natural images. Wang et al. \cite{wang2022perceptually} further achieved lossless compression of SCI through visibility modeling and depth prediction.
 
Inspired by this, Yang et al. \cite{itsrn0} designed an implicit coding scheme suitable for screen images, achieving preliminary SCISR.  Pak et al. \cite{btc} introduced B-spline curves into LTE, significantly reducing image blurring caused by the difference between screen resolution and SCI resolution, thus improving the quality of SCISR.
Shen et al. \cite{itsrn} constructed a new SCISR upsampling structure using the implicit Transformer to achieve excellent performance. 
However, the above methods are confined to describing the spatial pixel arrangement of SCI, lacking an exploration of its intrinsic amplitude‑phase characteristics.

\subsection{Frequency Decoupling in Image Super-Resolution}

Frequency modeling has become an effective paradigm in SR because it explicitly characterizes structural and textural information through the decomposition of image content into amplitude and phase components \cite{tmm5}.
Compared with spatial representations, amplitude and phase encode complementary properties of images. Amplitude mainly reflects global energy distribution, whereas phase preserves crucial structural organization. Such inherent decoupling provides a natural foundation for interpretable and structured representation learning in SR tasks. Early studies introduced frequency priors into image enhancement to improve visual reconstruction quality \cite{ffc}, \cite{craft}, \cite{fasa}. Building upon this perspective, Wang et al. \cite{sfmnet} emphasized decoupled frequency representations to separately characterize heterogeneous spectral components, while Wang et al. \cite{oct} further explored the complementary axial properties between amplitude and phase for complex-valued representation learning.
Meanwhile, Shi et al. \cite{decouple1} modeled image patches as waveform-based amplitude-phase representations, highlighting the importance of capturing representative periodic variations from spectral signals. Shah et al. \cite{decouple2} exploited frequency-decoupled interactions to preserve globally coherent spectral dependencies, demonstrating the effectiveness of context-aware frequency modeling. Becker et al. \cite{Thera} further revealed that adaptive coordination between amplitude attenuation and phase-preserving representations facilitates accurate structural correspondence and frequency-aware signal matching.
These studies collectively reveal that effective frequency modeling relies on proper decoupling, effective extraction, and faithful fitting.

Based on above analysis on SCI frequency characteristics, we decouple amplitude and phase representations to separately extract periodic pattern and coherent context, drawing on  aggregation strategies~\cite{CATA} that organize sparse high-energy patterns into compact prototypes and progressive attention mechanisms~\cite{PFA} that enhance focus without disrupting configurational continuity. These representations are further integrated through a unified implicit fitting scheme inspired by Fourier analysis networks~\cite{fan}, which explicitly models periodic and aperiodic representations for high-quality SCISR reconstruction.


	\section{Methodology}
\label{sec:meth}
We propose a SCISR framework that decouples amplitude and phase representations to separately capture periodic pattern and coherent context, further integrates these characteristics within an implicit fitting scheme for effective reconstruction.

\subsection{Overall Pipeline of Framework}
\begin{figure*}[t]
	\centering
	\begin{subfigure}{1\linewidth}
		\centering
		\includegraphics[width=1\linewidth]{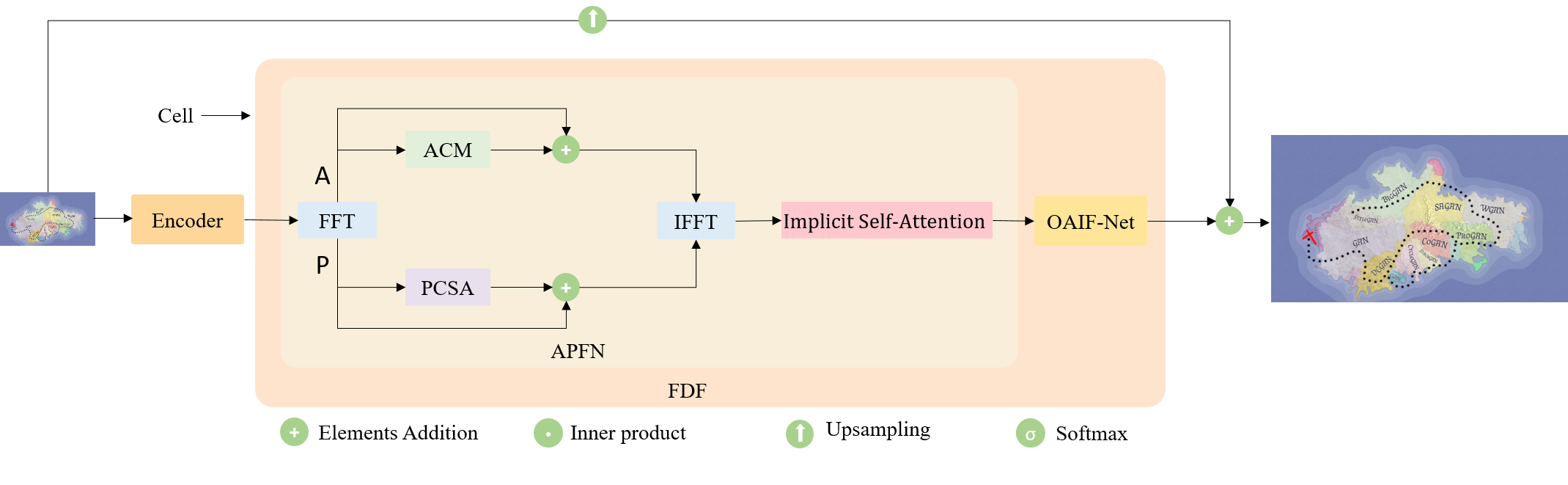}
	\end{subfigure}
	\caption{Overall architecture for SCISR.}
	\label{overall}
\end{figure*}

The overall architecture of FDF is illustrated in Fig.~\ref{overall}, which consists of an encoder, an APFN, and an OAIF-Net.

Given a LR image $I_{LR}\in\mathbb{R}^{H\times W\times 3}$, the encoder $E(\cdot)$ first extracts spatial-domain representations, and a Feature Enhancement Module~\cite{amsit} further enriches local contextual information for subsequent frequency-domain modeling. Instead of directly performing feature learning in the spatial domain, FDF reformulates SCI reconstruction from a frequency-decoupled perspective by transforming the enhanced features into the frequency domain through FFT, yielding $\mathcal{Z}_{in}=(\mathcal{A},\mathcal{P})$, where $\mathcal{A}$ and $\mathcal{P}$ denote amplitude and phase components, respectively.
Within the proposed Amplitude-Phase Factorization Network (APFN), amplitude and phase are processed through two specialized branches according to their distinct characteristics. ACM focuses on the sparse yet structured high-energy distributions in amplitude representations and organizes representative spectral patterns through learnable clustering prototypes, facilitating structured modeling of periodic pattern. In parallel, PCSA leverages the configurational continuity preserved in phase representations by progressively reinforcing configuration-consistent dependencies, enabling more coherent contextual modeling. Residual connections are introduced in both branches to preserve the original frequency information during decoupled representation learning. After bespoke modeling, the processed amplitude and phase features are recombined and transformed back to the spatial domain through IFFT:
\begin{equation}
	\mathcal{Z}_{in}^{\prime}
	= (\mathrm{ACM}(\mathcal{A})+\mathcal{A},\;
	\mathrm{PCSA}(\mathcal{P})+\mathcal{P})
\end{equation}
The decoupled frequency representations are then integrated through implicit continuous modeling rather than discrete upsampling. Implicit Self-Attention establishes continuous correspondences between HR query coordinates and LR feature representations, enabling flexible feature aggregation across arbitrary scales. Finally, the fused implicit representation $\mathcal{Z}_{APFN}$ is fed into OAIF-Net, which jointly models periodic and coherent implicit representations within a unified fitting scheme to reconstruct both repetitive textures and global contextual configurations in SCI.The final HR image is obtained as:
\begin{equation}
	I_{HR}
	=
	\mathrm{OAIF\text{-}Net}
	\big(\mathbf{APFN}(\mathbf{E}_{\psi}(I_{LR}),\, Cell,\, \hat{\theta}^h\big)\big)
	+
	I^{\uparrow}_{LR}
\end{equation}
Here, $\theta^{h}$ denotes the queried HR coordinates in the continuous image space, while $Cell$ represents the spatial extent of each query pixel.

\subsection{Amplitude-Phase Factorization Network}

\subsubsection{Amplitude Clustering Module}

The amplitude spectrum of SCI shown in Fig. \ref{SCI1} typically exhibits regularly arranged high‑energy stripes along with sharp peaks and reveals that structurally meaningful information concentrates on a few representative energy patterns rather than spreading uniformly.
However, conventional convolution inevitably mixes high-energy responses with low-energy regions, resulting in noise covering this discrete and structured pattern. 
Liu et al.~\cite{CATA} proposed Content-Aware Token Aggregation, which organizes information into a compact set of shared prototypes based on content similarity and effectively captures repetitive patterns while improving feature compactness. Furthermore, Zeng et al.~\cite{DTCT} demonstrated that emphasis on informative tokens can lead to discriminative prototype learning.

Based on the above observations, we propose an Amplitude Clustering Module (ACM) to perform clustering-based periodic pattern modeling in the amplitude space. ACM reorganizes dispersed amplitude representations according to their information similarity, enabling representative high-energy patterns to emerge through global interaction and aggregation. By progressively grouping amplitude elements with similar spectral characteristics into shared representative prototypes, ACM establishes a more compact and structured representation space for periodic information modeling. Such a clustering-based paradigm allows dominant periodic responses to be emphasized, thereby facilitating more stable and discriminative extraction of periodic patterns.
ACM first applies a rearrangement operation to flatten the 2D feature map into a label sequence, thereby achieving unified modeling in the label space and using Amplitude Prototype Modulation for subsequent clustering.
\begin{figure}[t]
	\centering
	\begin{subfigure}{1\linewidth}
		\centering
		\includegraphics[width=1\linewidth]{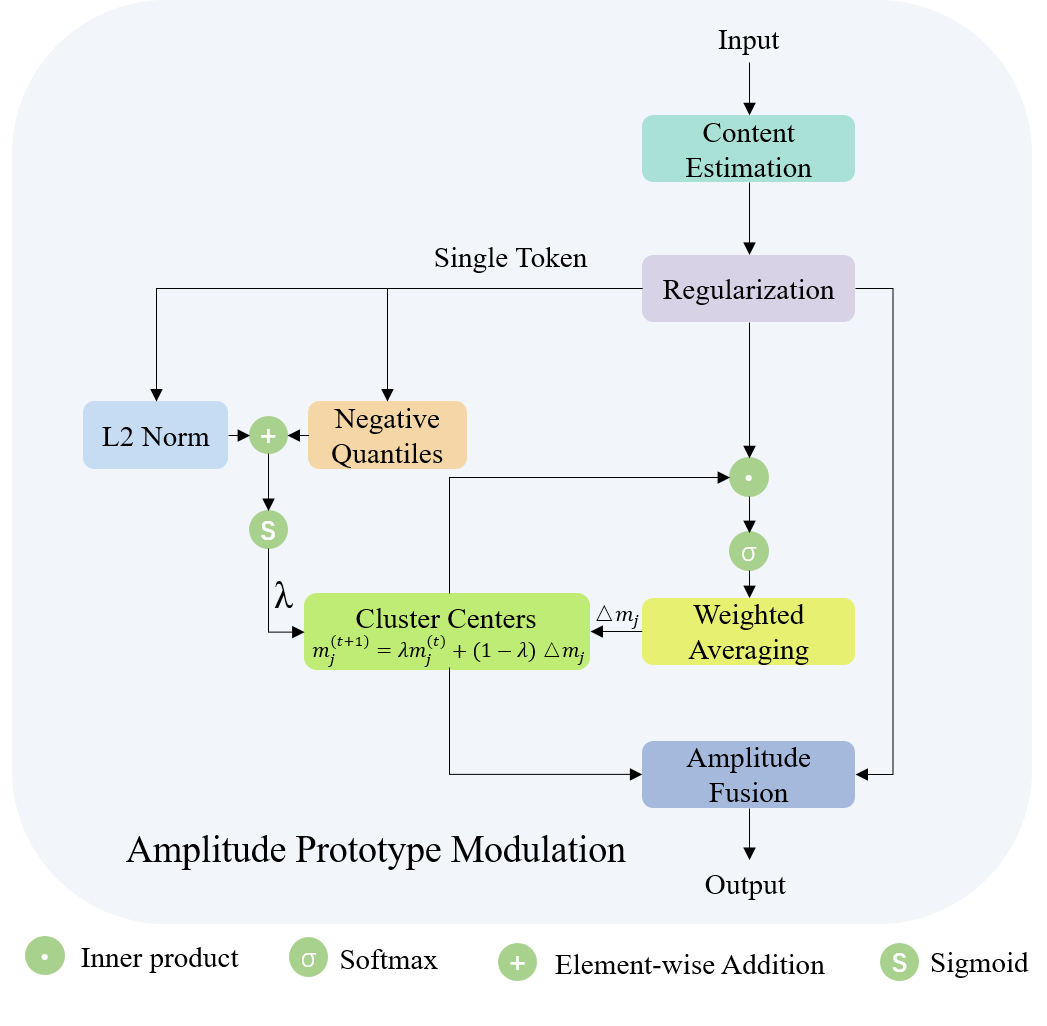}
	\end{subfigure}
	\caption{Schematic diagram of Amplitude Prototype Modulation.}
	\label{ACM}
\end{figure}

The schematic illustration of the Amplitude Prototype Modulation is shown in Fig.~\ref{ACM}. 
During clustering, ACM maintains a set of shared amplitude cluster centers.
The cluster centers are updated via weighted averaging:
\begin{equation}
	\frac{\sum_{i} p_{i,j} \mathbf{x}_i}{\sum_{i} p_{i,j}} \rightarrow
	\mathbf{m}_j 
\end{equation}
where $\mathbf{x}_i \in \mathbb{R}^{C}$ denotes the $i$-th amplitude token, $p_{i,j}$ denotes the soft assignment, $\hat{\mathbf{x}}_i$ is its normalized representation, $\mathbf{m}_j$ is the $j$-th cluster center, $\hat{\mathbf{m}}_j$ is its normalized form, $\langle\cdot,\cdot\rangle$ denotes inner-product similarity, and $M$ is the number of clusters. This process can be interpreted as soft grouping of tokens based on content similarity, enabling each cluster center to progressively capture a representative amplitude energy pattern.
To stabilize training, the cluster centers are further updated using an exponential moving average strategy, which mitigates drastic fluctuations in early training stages. Considering the significant variation in information content across amplitude tokens, ACM further introduces an amplitude-aware gating mechanism. For each token, a gating weight is defined as:
\begin{equation}
	g_i=S\left(\|\mathbf{x}_i\|_2-\mathrm{Quantile}\left(\left\{\|\mathbf{x}_k\|_2\right\}_{k=1}^{N},q\right)\right)
\end{equation}
where $S(\cdot)$ denotes the Sigmoid function, q represents its quantile proportion. This gating term adaptively modulates the contribution of clustered features, allowing high-energy tokens to receive stronger structural enhancement while preserving the original representation for low-energy tokens. 
Specifically, the exponential moving average provides a momentum coefficient $\lambda$ to stabilize the update of cluster centers, while the weighted aggregation term derived from soft assignments produces the current update direction $\Delta \mathbf{m}_j$. Formally, $\lambda \mathbf{m}_j^{(t)}$ preserves historical memory of the cluster center, and $(1-\lambda)\Delta \mathbf{m}_j$ incorporates the newly aggregated token information. Therefore, the update process of cluster centers can be formulated as:
\begin{equation}
	\mathbf{m}_j^{(t+1)} = \lambda \mathbf{m}_j^{(t)} + (1-\lambda)\Delta \mathbf{m}_j
\end{equation}
where $\mathbf{m}_j^{(t)}$ and $\mathbf{m}_j^{(t+1)}$ denote the cluster center of the $j$-th cluster at iteration $t$ and $t+1$, respectively, and $\Delta \mathbf{m}_j$ is computed from the assignment-weighted aggregation of tokens.
After clustering and feature aggregation, the tokens are reshaped back to the 2D feature map and further processed by convolutional layers with nonlinear activation.
At inference, ACM hard-assigns each token to its top-$K$ most relevant cluster centers. The value of $K$ aligns with the effective clustering range learned from the soft assignments during training, thereby ensuring numerical stability and structural consistency.

\begin{figure*}[t]
	\centering
	\begin{subfigure}{0.9\linewidth}
		\centering
		\includegraphics[width=1\linewidth]{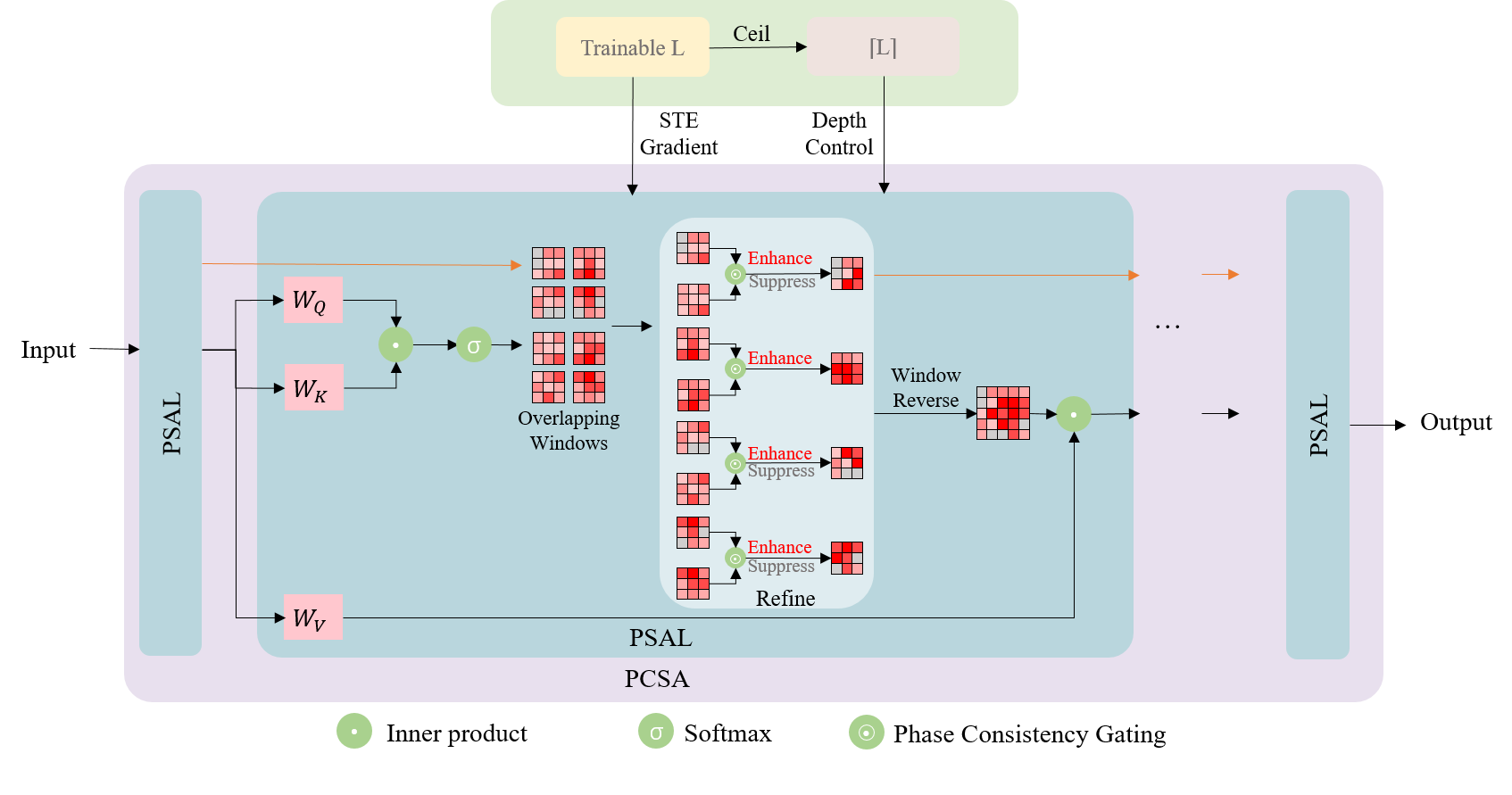}
	\end{subfigure}
	\caption{Schematic diagram of PCSA.}
	\label{PCSA}
\end{figure*}

\subsubsection{Phase Consistency Self-Attention}


Modeling coherent context from phase representations remains a challenging problem in SCI super-resolution. Since phase features encode delicate structural organizations, meaningful phase dependencies can be easily corrupted by noisy responses and inconsistent contextual interactions during feature propagation. Although Transformer-based methods~\cite{attention} provide powerful long-range modeling capabilities, they do not explicitly preserve structurally stable phase relationships, causing informative phase patterns to gradually weaken when mixed with irrelevant responses. Long et al.~\cite{PFA} proposed Progressive Focused Attention (PFA), which progressively suppresses less informative responses and highlights salient representations, demonstrating the effectiveness of progressive refinement for attention modeling. Building upon this insight, we propose a progressive consistency-driven mechanism for SCI phase modeling. It preserves structurally correlated phase patterns while avoiding the disruption caused by aggressive sparsification. Specifically, we propose Phase Consistency Self-Attention (PCSA), which continuously reinforces stable and configuration-consistent phase dependencies throughout attention propagation. By progressively strengthening structurally reliable relationships across different stages of feature interaction, PCSA enables meaningful phase patterns to gradually dominate contextual modeling while noise-induced responses naturally decay. Such consistency-aware refinement preserves the integrity and continuity of phase structures, facilitating effective extraction of coherent context in SCI.

The overall architecture of PCSA is illustrated in Figure~\ref{PCSA}. PCSA consists of multiple cascaded Phase Self-Attention Layers (PSALs), where the number of stacked layers is controlled by a trainable continuous parameter $L$ and discretized as $\lceil L \rceil$ during forward propagation. During backpropagation, gradients are approximated using straight-through estimator to enable optimization of the continuous depth parameter. This enables the network to dynamically balance representation capacity and computational complexity according to the characteristics of phase features. Except for the first layer, which only receives phase features, and the last layer, which only outputs refined phase features, intermediate PSALs simultaneously propagate both feature representations and attention maps across layers. Specifically, adjacent PSALs are coupled through cross-layer consistency propagation. Through this process, configuration-consistent responses are continuously enhanced, whereas unstable or noise-sensitive activations are gradually suppressed, enabling PCSA to effectively capture coherent context embedded in phase representations. In Fig.~\ref{PCSA}, black arrows denote the forward propagation of phase features and attention computation, while orange arrows represent the cross-layer transmission of attention maps, explicitly modeling the evolution of phase consistency. Given the input phase feature at the $l$-th PSAL, denoted as $\mathcal{P}^{(l)}$, the query, key, and value embeddings are first obtained through linear projections:
\begin{equation}
	Q^{(l)} = W_Q \mathcal{P}^{(l)}, \quad
	K^{(l)} = W_K \mathcal{P}^{(l)}, \quad
	V^{(l)} = W_V \mathcal{P}^{(l)},
\end{equation}
where $W_Q$, $W_K$, and $W_V$ denote learnable projection matrices. Under the local window constraint, the phase self-attention map is computed as:
\begin{equation}
	A^{(l)} = \mathrm{Softmax}\!\left(
	\frac{Q^{(l)} {K^{(l)}}^{T}}{\sqrt{d}} + R
	\right),
\end{equation}
where $d$ denotes the feature dimension and $R$ represents the relative positional encoding used to model the geometric relationship among phase variations. To explicitly enforce cross-layer phase consistency, the attention map from the previous layer $A^{(l-1)}$ is incorporated and fused with the current attention map via the Hadamard product:
\begin{equation}
	\tilde{A}^{(l)} = A^{(l)} \odot A^{(l-1)}.
\end{equation}
For the first PSAL, $A^{(l)}$ = $A^{(l-1)}$. This operation serves as a cross-layer consistency filtering mechanism, where only phase relationships that consistently exhibit strong responses across adjacent layers are preserved and amplified, while unstable or noise-driven correlations are progressively suppressed during propagation. The refined attention map is further propagated to subsequent PSALs for consistency refinement without performing spatial reconstruction, thereby preserving localized phase dependency modeling throughout the progressive refinement process. In contrast, the attention map used for value aggregation is restored to the original spatial arrangement through window reverse before interacting with $V^{(l)}$, ensuring that the refined phase representation maintains correct spatial correspondence in the image domain. Finally, the refined phase feature at the $l$-th layer is obtained by:
\begin{equation}
	\tilde{\mathcal{P}}^{(l)} = \tilde{A}^{(l)} V^{(l)}.
\end{equation}

\subsection{Oscillation-Anharmonic Implicit Fitting Network}

\begin{figure}[t]
	\centering
	\begin{subfigure}{0.75\linewidth}
		\centering
		\includegraphics[width=1\linewidth]{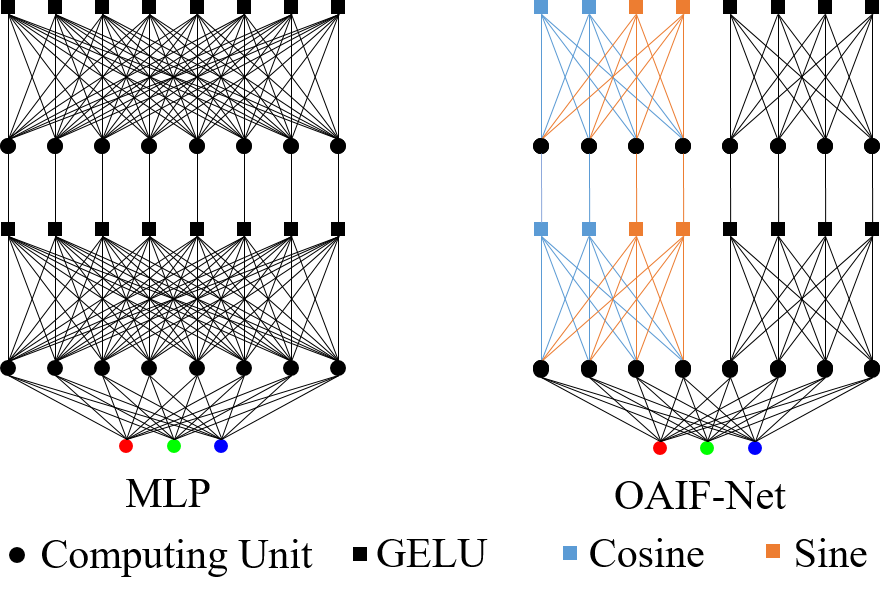}
	\end{subfigure}
	\caption{Structural comparison between MLP and OAIF-Net.}
	\label{FAN}
\end{figure}

The MLP-based methods are often inadequate for characterizing the periodic textures, repetitive structures, and sharp configurational transitions commonly observed in SCI. Liu et al.~\cite{liu2024finer} proposed variable-periodic activation functions to improve frequency adaptability and enhance the representation of high-frequency structures. Building upon this observation, Dong et al.~\cite{fan} further pointed out that existing MLPs and Transformers tend to memorize periodic data rather than genuinely model periodicity, and proposed Fourier Analysis Networks (FAN) to explicitly characterize periodic and aperiodic representations through trigonometric decomposition. Inspired by these advances, we seek to exploit the strong periodicity inherent in SCI and design a dedicated representation mechanism for effective modeling of repetitive structures and configurational transitions.
Herein, we further propose the Oscillation-Anharmonic Implicit Fitting Network (OAIF-Net). As illustrated in Fig.~\ref{FAN}, OAIF-Net treats periodic patterns and coherent context as equipollent representations, enabling repetitive structural patterns to be explicitly strengthened while maintaining globally consistent implicit contextual continuity. Such a unified fitting framework provides stable and expressive implicit representations for SCI and thus leads to faithful reconstruction of repetitive layouts, sharp configurational transitions, and coherent global context under arbitrary scaling factors.

The overall architecture of OAIF-Net consists of four periodically-enhanced implicit fitting layers followed by a linear output layer. Given an input continuous coordinate or feature vector $\mathbf{x} \in \mathbb{R}^{d}$, the mapping of the $l$-th periodically-enhanced layer is defined as:
\begin{equation}
	\mathbf{h}^{(l)} = \Big[ \cos(\mathbf{W}_p^{(l)} \mathbf{h}^{(l-1)}),\;
	\sin(\mathbf{W}_p^{(l)} \mathbf{h}^{(l-1)}),\;
	\phi(\mathbf{W}_g^{(l)} \mathbf{h}^{(l-1)}) \Big],
\end{equation}
where $\mathbf{h}^{(0)} = \mathbf{x}$, $\mathbf{W}_p^{(l)}$ and $\mathbf{W}_g^{(l)}$ denote the linear projection matrices for the periodic and non-periodic branches, respectively, and $\phi(\cdot)$ represents a nonlinear activation function such as GELU. By introducing both periodic and non-periodic branches within each layer, the network is able to simultaneously model structured periodic components and smooth non-periodic variations.

After multiple layers of periodically-enhanced mappings, OAIF-Net employs a standard linear projection as the output layer:
\begin{equation}
	\mathbf{y} = \mathbf{W}_{o} \mathbf{h}^{(L)} + \mathbf{b}_{o},
\end{equation}
where $L$ denotes the total number of periodically-enhanced layers. This design avoids introducing additional oscillations at the output stage, thereby improving numerical stability and reconstruction accuracy in the implicit fitting process.

\section{EXPERIMENTS}
\label{sec:expe}

\subsection{Implementation Details}

\subsubsection{Datasets and Metrics}

During training, we employ the SCI1K dataset~\cite{itsrn0}, a high-quality dataset specifically designed for screen content images. SCI1K contains 800 training images with a resolution of 720P, covering diverse application scenarios such as web pages, gaming scenes, cartoons, presentations, and documents. These images exhibit rich structural edges and high-contrast texture patterns. In addition, SCI1K provides 200 validation images with resolutions ranging from 720P to 2K, which are used to evaluate the reconstruction performance under different resolution settings.

For testing, we conduct comprehensive evaluations on multiple public benchmarks, including the SCI1K validation set \cite{itsrn0}, CCT~\cite{cct}, SCID~\cite{scid}, and SIQAD~\cite{siqad}. The CCT dataset is widely used for cross-content image quality assessment, containing both natural and screen images. In our experiments, we select 72 reference images excluding natural images for evaluation. The SCID dataset consists of 40 screen content images with 720P resolution, while the SIQAD dataset contains 20 images with a resolution of $600 \times 800$. Both datasets are commonly used in screen content image processing and quality assessment tasks.

Following standard practice, Peak Signal-to-Noise Ratio (PSNR) is adopted as the quantitative evaluation metric to measure the pixel-wise reconstruction accuracy between the super-resolved results and the corresponding high-resolution ground truth images.

\subsubsection{Training Settings}

The overall training pipeline follows common settings in screen image super-resolution tasks. Specifically, high-resolution images are first cropped into patches of size $48\eta \times 48\eta$, where $\eta$ denotes the upscaling factor. The corresponding low-resolution inputs are then generated by applying bicubic interpolation~\cite{pytorch} to downsample the cropped high-resolution patches.

To improve generalization ability, standard data augmentation techniques are applied to the low-resolution patches, including random horizontal flipping, vertical flipping, and $90^\circ$ rotations. Furthermore, we adopt re-parameterization and accumulation training strategies~\cite{amsit} to enhance robustness across different scaling factors.

From each high-resolution patch, $48^2$ pixels are randomly sampled, and their spatial coordinates along with the corresponding RGB values are used to form training pairs as supervision signals. The model is optimized using the Adam optimizer~\cite{Adam} with the $L_1$ loss function. The batch size is set to 16.

For learning rate scheduling, we employ a cosine annealing strategy~\cite{Cosine}. The model is trained for 1000 epochs, with an initial learning rate of $1 \times 10^{-5}$. A warm-up strategy is applied during the first 50 epochs, gradually increasing the learning rate to $1 \times 10^{-4}$, followed by a stable training phase.


\subsection{Comparison with State-of-the-Art Methods}

\subsubsection{Quantitative Analysis}

To comprehensively validate the effectiveness of the proposed FDF, we conduct quantitative comparisons with a range of representative and state-of-the-art methods. For fair and consistent evaluation, all compared methods adopt RDN~\cite{rdn} as a unified encoder. The comparison includes RDN, MetaSR~\cite{metasr}, LIIF~\cite{liif}, LTE~\cite{lte}, BTC~\cite{btc}, ITSRN++~\cite{itsrn}, and Thera~\cite{Thera}, which cover diverse technical paradigms in recent super-resolution research.

The quantitative results are summarized in Table~\ref{tab:SOTA}. Across multiple continuous and fixed scaling factors on four benchmark datasets, the proposed FDF consistently achieves the best performance. This advantage is not only reflected in objective metrics, but also demonstrates the superiority of FDF in feature modeling and scale generalization for screen image super-resolution. These results further verify its robustness and generalization capability across varying image contents and scaling factors.

\begin{table*}[t]
	\tiny
	\centering
	\resizebox{1\linewidth}{!}{
		\begin{tabular}{c|c|c|ccccccccc}
			\hline
			Dataset & Method & Params & $\times2$ & $\times3$ & $\times4$ & $\times5$ & $\times6$ & $\times7$ & $\times8$ & $\times9$ & $\times10$ \\
			\hline
			\multirow{8}{*}{SCI1K\cite{itsrn0}}
			& RDN\cite{rdn}      & 21.97M & 38.45 & 33.59 & 29.81 &  --  &  --  &  --  &  --  &  --  &  --  \\
			& MetaSR\cite{metasr}   & 22.42M & 38.57 & 33.67 & 30.12 & 27.52 & 26.13 & 23.91 & 23.19 & 22.02 & 21.73 \\
			& LIIF\cite{liif}     & 22.32M & 38.65 & 33.97 & 30.55 & 27.77 & 26.07 & 23.99 & 23.24 & 22.18 & 21.81 \\
			& LTE\cite{lte}      & 22.53M & 39.14 & 34.38 & 30.83 & 28.19 & 25.93 & 24.23 & 23.17 & 22.34 & 21.79 \\
			& BTC\cite{btc}      & 22.40M & 39.19 & 34.57 & 30.96 & 28.31 & 26.14 & 24.47 & 23.30 & 22.44 & 21.83 \\
			& ITSRN++\cite{itsrn}  & 23.95M & 39.26 & 34.71 & 31.08 & 28.45 & 26.21 & 24.55 & 23.37 & 22.53 & 21.91 \\
			& Thera\cite{Thera}    & 26.65M & 39.34 & 34.91 & 31.22 & 28.59 & 26.29 & 24.61 & 23.43 & 22.61 & 21.96 \\
			& FDF(Ours)      & 26.60M & \textbf{39.54} & \textbf{35.09} & \textbf{31.40} & \textbf{28.75} & \textbf{26.45} & \textbf{24.76} & \textbf{23.56} & \textbf{22.72} & \textbf{22.05} \\
			\hline
			\multirow{8}{*}{SCID\cite{scid}}
			& RDN\cite{rdn}      & 21.97M & 34.00 & 28.34 & 25.74 &  --  &  --  &  --  &  --  &  --  &  --  \\
			& MetaSR\cite{metasr}    & 22.42M & 33.84 & 29.08 & 25.76 & 23.62 & 22.38 & 21.59 & 21.07 & 20.71 & 20.41 \\
			& LIIF\cite{liif}     & 22.32M & 34.24 & 29.12 & 25.89 & 23.77 & 22.53 & 21.73 & 21.21 & 20.84 & 20.54 \\
			& LTE\cite{lte}      & 22.53M & 34.47 & 29.60 & 26.31 & 24.04 & 22.68 & 21.85 & 21.31 & 20.91 & 20.59 \\
			& BTC\cite{btc}      & 22.40M & 34.46 & 29.56 & 26.30 & 24.05 & 22.66 & 21.83 & 21.28 & 20.92 & 20.62 \\
			& ITSRN++\cite{itsrn}  & 23.95M & 34.55 & 29.67 & 26.42 & 24.15 & 22.73 & 21.87 & 21.35 & 20.98 & 20.68 \\
			& Thera\cite{Thera}    & 26.65M & 34.60 & 29.73 & 26.50 & 24.21 & 22.78 & 21.90 & 21.41 & 21.04 & 20.72 \\
			& FDF(Ours)      & 26.60M & \textbf{34.78} & \textbf{29.90} & \textbf{26.68} & \textbf{24.36} & \textbf{22.91} & \textbf{22.03} & \textbf{21.51} & \textbf{21.13} & \textbf{20.81} \\
			\hline
			\multirow{8}{*}{SIQAD\cite{siqad}}
			& RDN\cite{rdn}      & 21.97M & 33.53 & 26.89 & 23.38 &  --  &  --  &  --  &  --  &  --  &  --  \\
			& MetaSR\cite{metasr}   & 22.42M & 34.12 & 28.40 & 23.55 & 21.18 & 20.18 & 19.63 & 19.25 & 18.94 & 18.65 \\
			& LIIF\cite{liif}     & 22.32M & 34.31 & 28.27 & 23.44 & 21.16 & 20.25 & 19.70 & 19.36 & 19.02 & 18.70 \\
			& LTE\cite{lte}      & 22.53M & 35.01 & 29.35 & 24.23 & 21.54 & 20.40 & 19.78 & 19.42 & 19.09 & 18.83 \\
			& BTC\cite{btc}      & 22.40M & 35.01 & 29.40 & 24.25 & 21.56 & 20.44 & 19.81 & 19.43 & 19.12 & 18.82 \\
			& ITSRN++\cite{itsrn}  & 23.95M & 35.07 & 29.48 & 24.33 & 21.61 & 20.49 & 19.85 & 19.48 & 19.18 & 18.88 \\
			& Thera\cite{Thera}    & 26.65M & 35.13 & 29.54 & 24.40 & 21.65 & 20.55 & 19.89 & 19.53 & 19.23 & 18.92 \\
			& FDF(Ours)      & 26.60M & \textbf{35.32} & \textbf{29.70} & \textbf{24.57} & \textbf{21.78} & \textbf{20.69} & \textbf{20.01} & \textbf{19.63} & \textbf{19.32} & \textbf{18.99} \\
			\hline
			\multirow{8}{*}{CCT\cite{cct}}
			& RDN\cite{rdn}      & 21.97M & 33.62 & 30.28 & 28.34 &  --  &  --  &  --  &  --  &  --  &  --  \\
			& MetaSR\cite{metasr}   & 22.42M & 33.87 & 30.36 & 28.51 & 27.12 & 26.26 & 25.73 & 25.26 & 24.79 & 24.34 \\
			& LIIF\cite{liif}     & 22.32M & 34.22 & 30.65 & 28.57 & 27.18 & 26.29 & 25.76 & 25.23 & 24.97 & 24.70 \\
			& LTE\cite{lte}      & 22.53M & 34.47 & 30.75 & 28.63 & 27.24 & 26.35 & 25.81 & 25.35 & 25.01 & 24.63 \\
			& BTC\cite{btc}      & 22.40M & 34.55 & 30.79 & 28.65 & 27.32 & 26.47 & 25.85 & 25.39 & 25.04 & 24.69 \\
			& ITSRN++\cite{itsrn}  & 23.95M & 34.72 & 30.92 & 28.74 & 27.41 & 26.58 & 25.92 & 25.48 & 25.10 & 24.76 \\
			& Thera\cite{Thera}    & 26.65M & 34.88 & 31.05 & 28.82 & 27.49 & 26.65 & 25.94 & 25.54 & 25.14 & 24.81 \\
			& FDF(Ours)      & 26.60M & \textbf{35.06} & \textbf{31.21} & \textbf{28.99} & \textbf{27.64} & \textbf{26.79} & \textbf{26.07} & \textbf{25.66} & \textbf{25.23} & \textbf{24.90} \\
			\hline
		\end{tabular}
	}
	\caption{Quantitative comparisons with state-of-the-art methods on multiple datasets. The best-performing results are highlighted in $\textbf{Bold}$.}
	\label{tab:SOTA}
\end{table*}

\subsubsection{Qualitative Analysis}

\begin{figure*}[thbp]
	\centering
	\setlength{\tabcolsep}{3pt}
	\renewcommand{\arraystretch}{0.9}
	
	\begin{tabular}{cccc}
		& \makebox[0.25\linewidth][c]{SCI1K 00067($\times 6$)} & \makebox[0.25\linewidth][c]{SCID 00023($\times 4$)} & \makebox[0.25\linewidth][c]{SIQAD 00002($\times 2$)} \\
		
		\makebox[0.11\linewidth][l]{LIIF} &
		\raisebox{-0.5\height}{\includegraphics[width=0.25\linewidth]{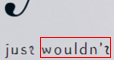}} &
		\raisebox{-0.5\height}{\includegraphics[width=0.25\linewidth]{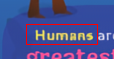}} &
		\raisebox{-0.5\height}{\includegraphics[width=0.25\linewidth]{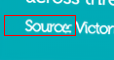}} \\
		& \makebox[0.25\linewidth][c]{wouldn'\textcolor{red}{7}} & \makebox[0.25\linewidth][c]{Hum\textcolor{red}{oR}} & \makebox[0.25\linewidth][c]{Sour\textcolor{red}{o}e} \\
		& \makebox[0.25\linewidth][c]{97.52} & \makebox[0.25\linewidth][c]{97.18} & \makebox[0.25\linewidth][c]{98.63} \\
		
		\makebox[0.11\linewidth][l]{LTE} &
		\raisebox{-0.5\height}{\includegraphics[width=0.25\linewidth]{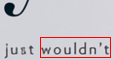}} &
		\raisebox{-0.5\height}{\includegraphics[width=0.25\linewidth]{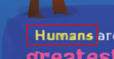}} &
		\raisebox{-0.5\height}{\includegraphics[width=0.25\linewidth]{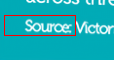}} \\
		& \makebox[0.25\linewidth][c]{wouldn't} & \makebox[0.25\linewidth][c]{Humans} & \makebox[0.25\linewidth][c]{Sour\textcolor{red}{o}e} \\
		& \makebox[0.25\linewidth][c]{99.87} & \makebox[0.25\linewidth][c]{98.15} & \makebox[0.25\linewidth][c]{99.81} \\
		
		\makebox[0.11\linewidth][l]{BTC} &
		\raisebox{-0.5\height}{\includegraphics[width=0.25\linewidth]{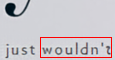}} &
		\raisebox{-0.5\height}{\includegraphics[width=0.25\linewidth]{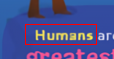}} &
		\raisebox{-0.5\height}{\includegraphics[width=0.25\linewidth]{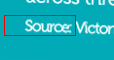}} \\
		& \makebox[0.25\linewidth][c]{wouldn'\textcolor{red}{7}} & \makebox[0.25\linewidth][c]{Hum\textcolor{red}{B}ns} & \makebox[0.25\linewidth][c]{Sour\textcolor{red}{o}e} \\
		& \makebox[0.25\linewidth][c]{97.73} & \makebox[0.25\linewidth][c]{97.85} & \makebox[0.25\linewidth][c]{99.80} \\
		
		\makebox[0.11\linewidth][l]{ITSRN++} &
		\raisebox{-0.5\height}{\includegraphics[width=0.25\linewidth]{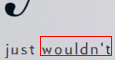}} &
		\raisebox{-0.5\height}{\includegraphics[width=0.25\linewidth]{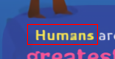}} &
		\raisebox{-0.5\height}{\includegraphics[width=0.25\linewidth]{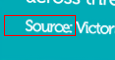}} \\
		& \makebox[0.25\linewidth][c]{wouldn't} & \makebox[0.25\linewidth][c]{Humans} & \makebox[0.25\linewidth][c]{Sour\textcolor{red}{o}e} \\
		& \makebox[0.25\linewidth][c]{99.91} & \makebox[0.25\linewidth][c]{98.47} & \makebox[0.25\linewidth][c]{99.82} \\
		
		\makebox[0.11\linewidth][l]{Thera} &
		\raisebox{-0.5\height}{\includegraphics[width=0.25\linewidth]{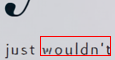}} &
		\raisebox{-0.5\height}{\includegraphics[width=0.25\linewidth]{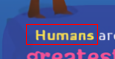}} &
		\raisebox{-0.5\height}{\includegraphics[width=0.25\linewidth]{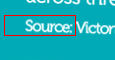}} \\
		& \makebox[0.25\linewidth][c]{wouldn't} & \makebox[0.25\linewidth][c]{Humans} & \makebox[0.25\linewidth][c]{Sour\textcolor{red}{o}e} \\
		& \makebox[0.25\linewidth][c]{99.85} & \makebox[0.25\linewidth][c]{98.49} & \makebox[0.25\linewidth][c]{99.81} \\
		
		\makebox[0.11\linewidth][l]{FDF(Ours)} &
		\raisebox{-0.5\height}{\includegraphics[width=0.25\linewidth]{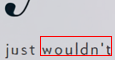}} &
		\raisebox{-0.5\height}{\includegraphics[width=0.25\linewidth]{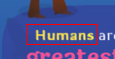}} &
		\raisebox{-0.5\height}{\includegraphics[width=0.25\linewidth]{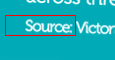}} \\
		& \makebox[0.25\linewidth][c]{wouldn't} & \makebox[0.25\linewidth][c]{Humans} & \makebox[0.25\linewidth][c]{Source} \\
		& \makebox[0.25\linewidth][c]{\textbf{99.94}} & \makebox[0.25\linewidth][c]{\textbf{98.69}} & \makebox[0.25\linewidth][c]{\textbf{99.87}} \\
		
		\makebox[0.11\linewidth][l]{GT} &
		\raisebox{-0.5\height}{\includegraphics[width=0.25\linewidth]{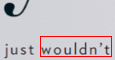}} &
		\raisebox{-0.5\height}{\includegraphics[width=0.25\linewidth]{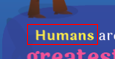}} &
		\raisebox{-0.5\height}{\includegraphics[width=0.25\linewidth]{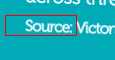}} \\
		& \makebox[0.25\linewidth][c]{wouldn't} & \makebox[0.25\linewidth][c]{Humans} & \makebox[0.25\linewidth][c]{Source} \\
		& \makebox[0.25\linewidth][c]{99.98} & \makebox[0.25\linewidth][c]{99.95} & \makebox[0.25\linewidth][c]{99.91} \\
	\end{tabular}
	
	\caption{
		Qualitative comparison of FDF with other methods at multiple integer magnification levels: We employed a pre-trained scene text recognition network (STR) to identify the red-boxed regions, yielding prediction results and average confidence scores. Incorrect predictions are marked in {\color{red}red}, while the highest average confidence score in the super-resolved image is highlighted in {\bfseries bold}.
	}
	\label{fig:integer}
\end{figure*}

To further provide intuitive insights into the visual reconstruction capability of FDF, we conduct qualitative comparisons on representative samples from SCI1K, SCID, and SIQAD datasets under different integer scaling factors. The results are shown in Fig.~\ref{fig:integer}. All methods employ RDN~\cite{rdn} as the encoder under a unified evaluation setting to ensure fairness.

Moreover, to comprehensively assess the reconstruction quality of textual structures, we incorporate auxiliary analysis using text recognition results. Since screen images often contain abundant textual content, their reconstruction quality can be indirectly reflected by recognition accuracy. Specifically, we employ a pre-trained Scene Text Recognition (STR) network~\cite{str} to recognize text within the red-boxed regions of the reconstructed images, and report both predicted text and average confidence scores as auxiliary evaluation metrics.

For the sample $00067$ from the SCI1K dataset under $\times 6$ scaling, LIIF and BTC exhibit noticeable structural distortions in character details, such as confusion between apostrophes and the digit “7,” leading to incorrect recognition results. Although LTE, ITSRN++, and Thera improve contour continuity to some extent, they still suffer from blurred edges and locally unstable structures. In contrast, FDF accurately reconstructs the word “wouldn't,” with clear edges and natural stroke connections, achieving the best visual quality.

For the sample $00023$ from the SCID dataset under $\times 4$ scaling, significant differences can be observed in reconstructing the word “Humans.” LIIF and BTC produce inconsistent internal structures, resulting in misrecognition of certain characters. ITSRN++ and Thera can recover the overall text shape but still exhibit inconsistencies in stroke thickness and continuity. FDF achieves the most accurate reconstruction, preserving both edge integrity and internal structural consistency.

For the sample $00002$ from the SIQAD dataset under $\times 2$ scaling, the reconstruction of the word “Source” also shows clear differences among methods. Several approaches introduce distortions or blurring in characters such as “o” and “e,” degrading readability. Notably, all compared methods except FDF misrecognize the letter “c” as “o.” In contrast, FDF is the only method that preserves the closed structures and edge continuity of characters, resulting in the clearest and most recognizable text. Across all experiments, FDF achieves the highest average confidence scores, indicating superior readability and reconstruction fidelity.

\begin{figure*}[thbp]
	\centering
	\setlength{\tabcolsep}{3pt}
	\renewcommand{\arraystretch}{1.2}
	
	\begin{tabular}{cccc}
		\makebox[0.2\linewidth][c]{SIQAD 00013 $\times$3.51} &
		\makebox[0.2\linewidth][c]{LIIF} &
		\makebox[0.2\linewidth][c]{LTE} &
		\makebox[0.2\linewidth][c]{BTC} \\
		
		
		\raisebox{-0.5\height}{\includegraphics[width=0.22\linewidth]{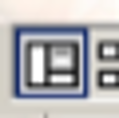}} &
		\raisebox{-0.5\height}{\includegraphics[width=0.22\linewidth]{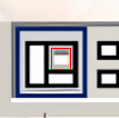}} &
		\raisebox{-0.5\height}{\includegraphics[width=0.22\linewidth]{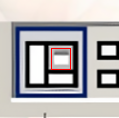}} &
		\raisebox{-0.5\height}{\includegraphics[width=0.22\linewidth]{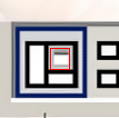}} \\
		
		
		\makebox[0.2\linewidth][c]{ITSRN++} &
		\makebox[0.2\linewidth][c]{Thera} &
		\makebox[0.2\linewidth][c]{FDF} &
		\makebox[0.2\linewidth][c]{GT} \\
		
		
		\raisebox{-0.5\height}{\includegraphics[width=0.22\linewidth]{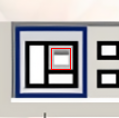}} &
		\raisebox{-0.5\height}{\includegraphics[width=0.22\linewidth]{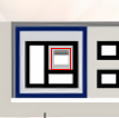}} &
		\raisebox{-0.5\height}{\includegraphics[width=0.22\linewidth]{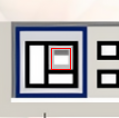}} &
		\raisebox{-0.5\height}{\includegraphics[width=0.22\linewidth]{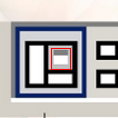}} \\
		

		\makebox[0.2\linewidth][c]{SCID 00031 $\times$2.43} &
		\makebox[0.2\linewidth][c]{LIIF} &
		\makebox[0.2\linewidth][c]{LTE} &
		\makebox[0.2\linewidth][c]{BTC} \\
		
		
		\raisebox{-0.5\height}{\includegraphics[width=0.22\linewidth]{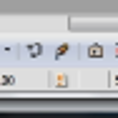}} &
		\raisebox{-0.5\height}{\includegraphics[width=0.22\linewidth]{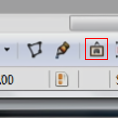}} &
		\raisebox{-0.5\height}{\includegraphics[width=0.22\linewidth]{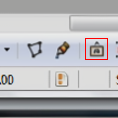}} &
		\raisebox{-0.5\height}{\includegraphics[width=0.22\linewidth]{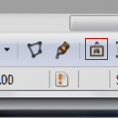}} \\
		
		
		\makebox[0.2\linewidth][c]{ITSRN++} &
		\makebox[0.2\linewidth][c]{Thera} &
		\makebox[0.2\linewidth][c]{FDF} &
		\makebox[0.2\linewidth][c]{GT} \\
		
		
		\raisebox{-0.5\height}{\includegraphics[width=0.22\linewidth]{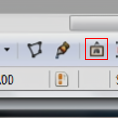}} &
		\raisebox{-0.5\height}{\includegraphics[width=0.22\linewidth]{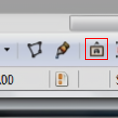}} &
		\raisebox{-0.5\height}{\includegraphics[width=0.22\linewidth]{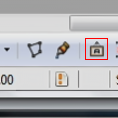}} &
		\raisebox{-0.5\height}{\includegraphics[width=0.22\linewidth]{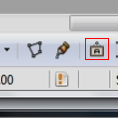}} \\
	\end{tabular}
	
	\vspace{2mm}
	
	\caption{
		Qualitative comparison of FDF and other methods at non-integer scales. The areas highlighted in red indicate regions of significant difference.
	}
	\label{fig:nointeger}
\end{figure*}

In addition, Fig.~\ref{fig:nointeger} presents qualitative comparisons under non-integer scaling factors. Using the same RDN encoder, we evaluate reconstruction stability and fine-detail recovery under continuous scaling conditions.

For the sample $00013$ from the SIQAD dataset with a scaling factor of $\times 3.51$, significant differences can be observed in the reconstruction of user interface icons. LIIF and BTC suffer from texture misalignment and blurring around icon edges, with discontinuous or uneven line thickness. LTE and ITSRN++ improve the overall structure but still introduce artifacts in fine-grained texture regions, leading to unclear internal structures. In contrast, FDF accurately reconstructs geometric boundaries and internal textures, producing well-organized structures with smooth transitions. Particularly in thin lines and internal details, FDF maintains superior continuity and sharpness, achieving results closest to the ground truth.

For the sample $00031$ from the SCID dataset under $\times 2.43$ scaling, the reconstruction of complex user interface elements further highlights the differences among methods. Specifically, the character “A” in the top-left tab icon and the orange exclamation mark below pose significant challenges. LIIF and BTC produce severe distortions and broken strokes, making the character unrecognizable. LTE and Thera can restore the general shape but still fall short in edge sharpness and color fidelity. In contrast, FDF accurately reconstructs the geometric structure of the letter “A” and preserves the color distribution and edge details of the orange exclamation mark, effectively avoiding texture aliasing and artifacts. Its reconstruction results demonstrate superior structural consistency and visual realism.

Overall, these results demonstrate that FDF achieves strong generalization and stable reconstruction performance across different datasets and scaling factors. Whether handling regular geometric structures in user interface elements or complex regions containing text and symbols, FDF effectively suppresses artifacts and accurately restores critical details. Furthermore, it significantly improves the stability and accuracy of downstream text recognition, highlighting its effectiveness and practical value in real-world screen image applications.

\subsection{Ablation Studies}

\subsubsection{Effectiveness of ACM}

To validate the effectiveness of the proposed Amplitude Clustering Module (ACM) within the FDF, we conduct systematic ablation experiments on the SCI1K test set. The quantitative results are reported in Table~\ref{tab:FDT}. To improve experimental efficiency, we slightly reduce the training iterations. We compare the performance of the model with and without ACM under identical settings. As shown in the table, incorporating ACM consistently improves PSNR across all scaling factors, with more pronounced gains at higher scales. These improvements reveal ACM effectively enhances the modeling and reconstruction of periodic patterns that become increasingly difficult to preserve under large-scale upsampling.

\begin{figure*}[thbp]
	\centering
	\setlength{\tabcolsep}{3pt}
	\renewcommand{\arraystretch}{1.2}
	
	\begin{tabular}{cccc}
		\raisebox{-0.5\height}{\includegraphics[width=0.3\linewidth]{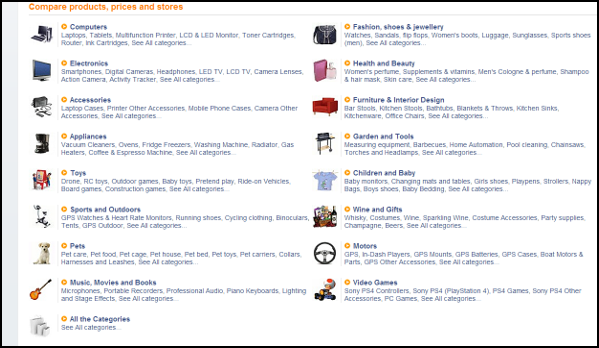}} &
		\raisebox{-0.5\height}{\includegraphics[width=0.3\linewidth]{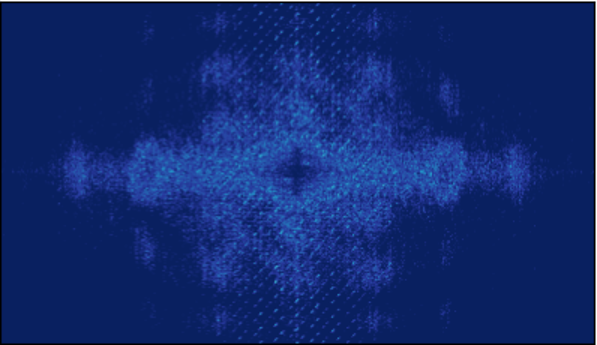}} &
		\raisebox{-0.5\height}{\includegraphics[width=0.3\linewidth]{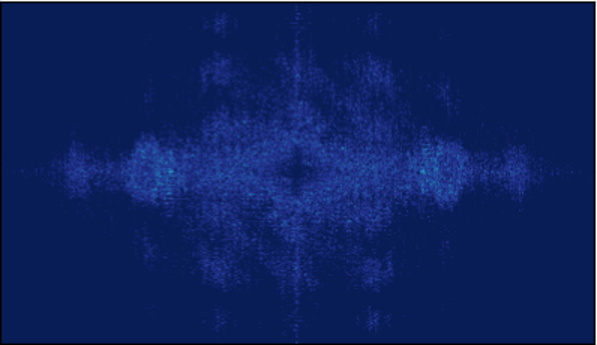}} &
		\hspace{-0.2em}\raisebox{-0.5\height}{\includegraphics[height=0.175\linewidth]{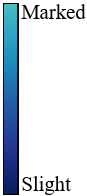}} 
		\\
		
		\makebox[0.24\linewidth][c]{SCID 00020 $\times$4} &
		\makebox[0.24\linewidth][c]{CATANet} &
		\makebox[0.24\linewidth][c]{ACM} &
	\end{tabular}
	
	\caption{
		Visual comparison of the ability of CATA and ACM to extract periodic pattern. The right side shows the indicator bars, with the error ranging from marked to slight, from bright to dark.
	}
	\label{fig:mem}
\end{figure*}

To further analyze the role of ACM from a frequency perspective, we visualize the amplitude spectral error between the SR results and the HR image, as shown in Fig. ~\ref{fig:mem}~\cite{mem}. Brighter regions correspond to larger amplitude reconstruction errors. Compared with the CATA method, ACM significantly suppresses the error intensity around the discrete spectral peaks and structured boundary regions, where periodic responses are densely concentrated. This demonstrates that ACM more effectively organizes sparse yet representative high-energy spectral patterns through amplitude prototype modulation, thereby strengthening the reconstruction of periodic structures and repetitive textures embedded in SCI while reducing overall amplitude distortion.

\subsubsection{Effectiveness of PCSA}



To evaluate the effectiveness of PCSA, we conducted ablation experiments comparing PFA and PCSA while keeping other components constant. The corresponding quantitative results are shown in Table~\ref{tab:FDT}. It can be seen that PCSA significantly improves performance under all scaling factors, indicating that phase consistency constraints have a positive impact on high-quality reconstruction.

\begin{figure}[thbp]
	\centering
	\setlength{\tabcolsep}{5pt}
	\renewcommand{\arraystretch}{1.0}
	
	\begin{tabular}{ccc}
		\includegraphics[width=0.3\linewidth]{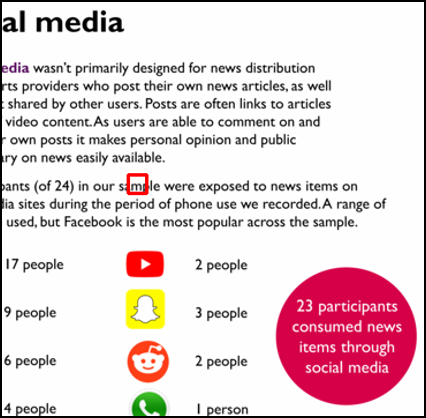} &
		\includegraphics[width=0.3\linewidth]{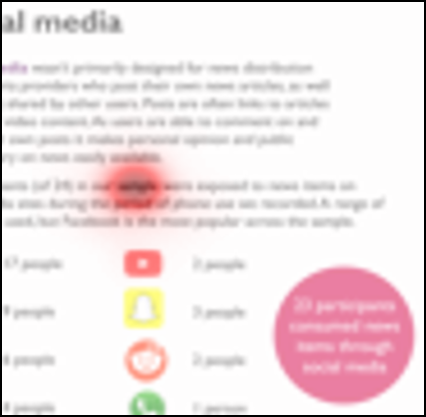} &
		\includegraphics[width=0.3\linewidth]{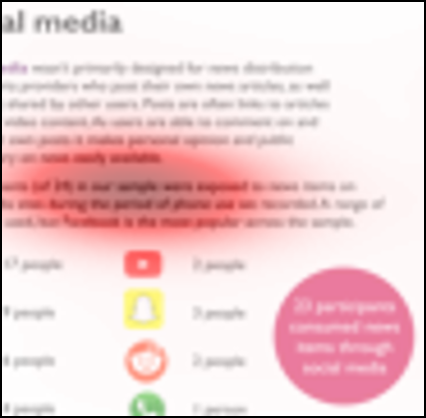} \\
		
		\makebox[0.3\linewidth][c]{\small SCI1K 00166 x4} &
		\makebox[0.3\linewidth][c]{\small PFA} &
		\makebox[0.3\linewidth][c]{\small PCSA} \\
	\end{tabular}
	
	
	\caption{
		Visualization Experiment on the Effect of PCSA on Receptive Fields.
	}
	\label{fig:lam}
\end{figure}

Furthermore, we used Local Attribution Map (LAM)~\cite{lam} to visualize the impact of PCSA on the receptive field of model. As shown in Fig.~\ref{fig:lam}, the red areas represent the regions that contribute the most to reconstruction; the higher the color intensity, the greater the contribution. Clearly, PCSA significantly expands the spatial coverage of high-response regions, especially around critical structural regions, indicating that PCSA effectively expands the receptive field.

\begin{table*}[!ht]
	\tiny
	\centering
	
	\resizebox{0.92\linewidth}{!}{
		\begin{tabular}{cc|cc|ccc|c|ccc}
			\hline
			\multicolumn{2}{c|}{Amplitude} 
			& \multicolumn{2}{c|}{Phase} 
			& \multicolumn{3}{c|}{Upsampling} 
			& \multirow{2}{*}{Params}
			& \multicolumn{3}{c}{SCI1K Test} \\
			
			\cline{1-7}\cline{9-11}
			
			CATA & ACM & PFA ~\cite{PFA} & PCSA
			& MLP\cite{liif} & FAN\cite{fan} & OAIF-Net 
			& 
			& $\times2$ & $\times4$ & $\times8$ \\
			\hline
			
			\checkmark &  & \checkmark &  
			& \checkmark &  &  
			& 27.36M
			& 39.02 & 30.69 & 23.02 \\
			
			& \checkmark & \checkmark &  
			& \checkmark &  &  
			& 27.34M
			& 39.15 & 30.82 & 23.14 \\
			
			\checkmark &  &  & \checkmark  
			& \checkmark &  &  
			& 27.39M
			& 39.21 & 30.93 & 23.19 \\
			
			& \checkmark &  & \checkmark  
			& \checkmark &  &  
			& 27.37M
			& 39.35 & 31.08 & 23.30 \\
			
			\checkmark &  & \checkmark &  
			&  & \checkmark &  
			& 27.02M
			& 39.12 & 30.80 & 23.11 \\
			
			& \checkmark & \checkmark &  
			&  & \checkmark &  
			& 27.00M
			& 39.24 & 30.94 & 23.22 \\
			
			\checkmark &  &  & \checkmark  
			&  & \checkmark &  
			& 27.05M
			& 39.31 & 31.02 & 23.28 \\
			
			& \checkmark &  & \checkmark  
			&  & \checkmark &  
			& 27.03M
			& 39.42 & 31.18 & 23.37 \\
			
			\checkmark &  & \checkmark &  
			&  &  & \checkmark 
			& 26.59M
			& 39.28 & 30.97 & 23.23 \\
			
			& \checkmark & \checkmark &  
			&  &  & \checkmark 
			& \textbf{26.57M}
			& 39.39 & 31.11 & 23.33 \\
			
			\checkmark &  &  & \checkmark  
			&  &  & \checkmark 
			& 26.62M
			& 39.46 & 31.22 & 23.40 \\
			
			& \checkmark &  & \checkmark  
			&  &  & \checkmark 
			& 26.60M
			& \textbf{39.58} & \textbf{31.36} & \textbf{23.53} \\
			
			\hline
		\end{tabular}
	}
	\caption{Results of the quantitative effectiveness experiments of FDF conducted on the SCI1K test set. The best-performing results are highlighted in $\textbf{Bold}$.}
	\label{tab:FDT}
\end{table*}

Through its progressive self-attention mechanism, PCSA explicitly introduces phase consistency constraints into the attention computation process. This encourages the model to focus on regions with similar phase structures during feature aggregation, rather than rely solely on local similarity. Since phase information inherently exhibits global correlations across spatial locations in the frequency domain, the phase-consistency-based attention helps establish more meaningful long-range dependencies, thereby enhancing the modeling of global structures and semantic coherence.

\subsubsection{Effectiveness of OAIF-Net}

\begin{figure*}[thbp]
	\centering
	\setlength{\tabcolsep}{4pt}
	\renewcommand{\arraystretch}{1.0}
	
	\begin{tabular}{cccc}
		\includegraphics[width=0.23\linewidth]{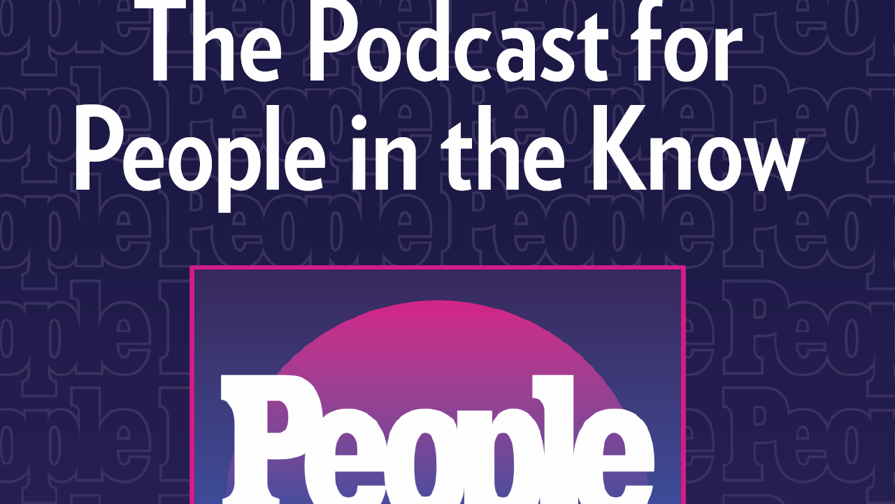} &
		\includegraphics[width=0.23\linewidth]{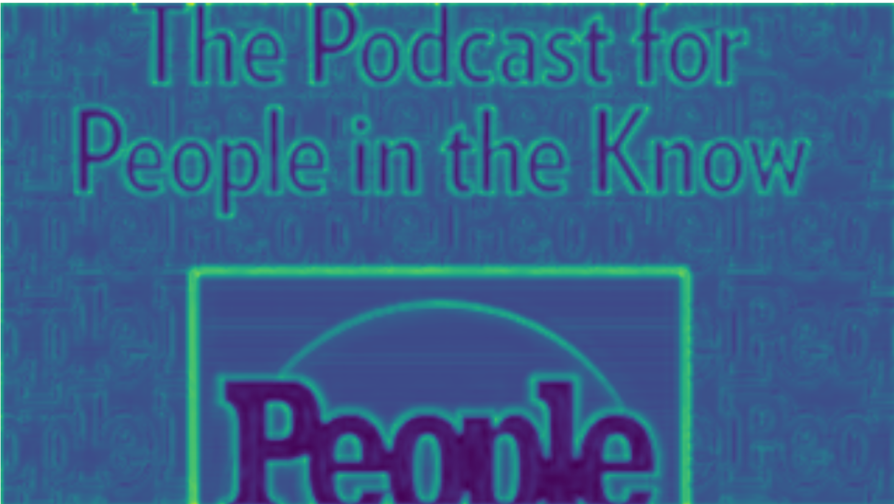} &
		\includegraphics[width=0.23\linewidth]{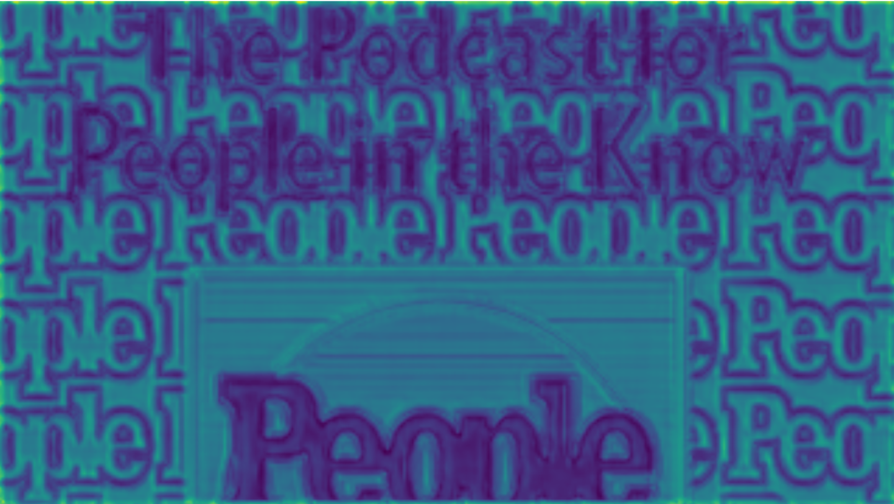} &
		\includegraphics[width=0.23\linewidth]{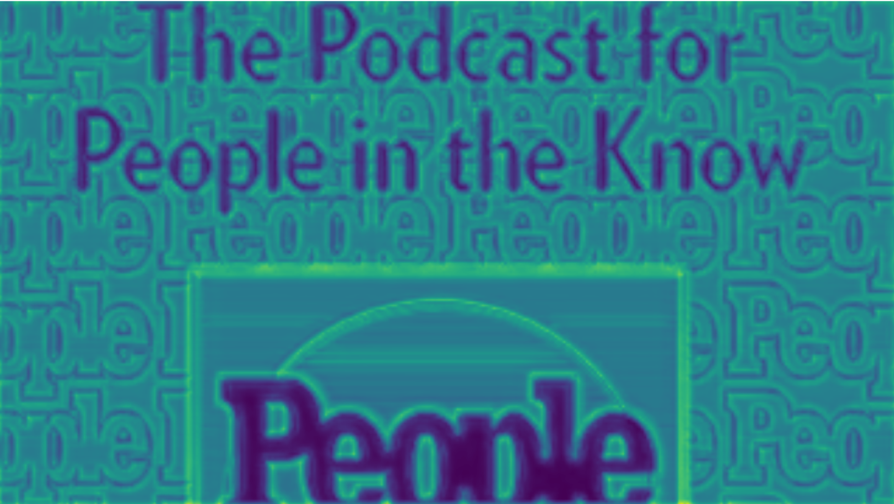} \\
		
		\makebox[0.23\linewidth][c]{\small SCI1K 00065 x4} &
		\makebox[0.23\linewidth][c]{\small MLP} &
		\makebox[0.23\linewidth][c]{\small FAN} &
		\makebox[0.23\linewidth][c]{\small OAIF-Net} \\
	\end{tabular}
	
	
	\caption{
		Visual feature maps obtained from elements extracted in the connection layer of different upsampling structures.
	}
	\label{fig:vis}
\end{figure*}

OAIF-Net improves the modeling of repetitive patterns by explicitly introducing periodically-enhanced mappings into the hidden layers. Its multi-layer periodic enhancement structure enables progressive combination and evolution of periodic features, making it more suitable for implicit representation of complex screen images. As shown in Table~\ref{tab:FDT}, OAIF-Net achieves superior reconstruction performance compared to existing upsampling methods across all scaling factors, demonstrating its stronger representation capability and structural modeling advantage in continuous-scale screen image super-resolution.

To further analyze the differences among implicit upsampling structures, we visualize intermediate feature representations from three models, namely MLP, FAN, and the proposed OAIF-Net, as shown in Fig.~\ref{fig:vis}, where green regions indicate high feature response intensity corresponding to areas that receive greater network attention. It can be observed that all methods effectively suppress redundant responses in flat white regions after backbone feature extraction, suggesting that the extracted features already possess a certain degree of discriminability. However, when a standard MLP is employed as the implicit fitting function, feature responses are primarily concentrated around character boundaries, with limited activation in background and structural regions, resulting in insufficient modeling of complex textures and non-uniform backgrounds. In contrast, FAN explicitly models periodic patterns and exhibits strong capability in capturing repetitive structures. Nevertheless, its emphasis on periodicity introduces a response bias, leading to excessive attention to repetitive textures while weakening the distinction of salient foreground text, thereby reducing the discriminability of the response distribution. Compared with these methods, OAIF-Net achieves a more balanced feature representation by integrating periodic enhancement with implicit fitting. It preserves sensitivity to repetitive structures without overemphasizing specific texture patterns, enabling more informative responses across both foreground and background regions. Consequently, OAIF-Net produces clearer response distributions and stronger structural discrimination, yielding the most informative and well-separated feature representations among all compared methods.

\subsection{Limitations}

Although the proposed FDF achieves strong performance in arbitrary-scale SCI super-resolution, several aspects remain worthy of further exploration. Future work may focus on designing periodicity-guided positional encodings to better exploit repetitive structural priors in SCI, as well as developing frequency-adaptive routing mechanisms that dynamically allocate different periodic patterns, such as text edges and sub-pixel layouts. In addition, constructing lightweight variants through parameter sharing or efficient implicit fitting strategies could further improve inference efficiency and facilitate deployment in real-world resource-constrained scenarios.






	\section{Conclusion}
	
	In this work, we investigate the intrinsic amplitude–phase characteristics of screen content images (SCI) for arbitrary-scale super-resolution and propose a Frequency Decoupled Framework (FDF). Through systematic analysis, SCI exhibit highly heterogeneous amplitude–phase characteristics. Amplitude representations mainly encode sparse yet structured periodic energy aggregation, while phase representations predominantly preserve coherent structural continuity and global contextual dependencies.
	Built upon this insight, we propose a Frequency Decoupled Framework (FDF) consisting of an Amplitude–Phase Factorization Network (APFN) and an Oscillation-Anharmonic Implicit Fitting Network (OAIF-Net). Amplitude Clustering Module (ACM) in APFN organizes sparse high-energy spectral responses into representative prototypes to enhance periodic pattern extraction, while the Phase Consistency Self-Attention (PCSA) in APFN progressively reinforces configurationally phase dependencies through consistency-aware contextual propagation to capture coherent context. OAIF-Net further integrates periodic pattern and coherent context within a unified implicit fitting scheme to effectively exploit the aforementioned information, thereby achieving faithful reconstruction of SCI at arbitrary scales.
	Extensive experiments across multiple benchmarks demonstrate FDF achieves consistent and significant improvements over state-of-the-art methods under mulyiple scaling factors. These results verify both periodic patterns and coherent context should be effectively captured and exploited to achieve high-quality SCI super-resolution.
	In the future, we will explore periodicity-guided positional encoding, frequency-adaptive interaction, and efficient implicit fitting schemes for light and versatile arbitrary-scale SCI reconstruction.

	{\small
		\bibliographystyle{unsrt}
		\bibliography{egbib.bib}
	}
	
	\vspace{-20mm}

	\vspace{-10mm}


\end{document}